\documentclass[runningheads]{llncs}
\usepackage{graphicx}
\usepackage{hyperref}
\usepackage{listings}
\usepackage{xcolor}

\definecolor{codegreen}{rgb}{0,0.6,0}
\definecolor{codegray}{rgb}{0.5,0.5,0.5}
\definecolor{codepurple}{rgb}{0.58,0,0.82}
\definecolor{backcolour}{rgb}{0.95,0.95,0.92}

\lstdefinestyle{mystyle}{
    backgroundcolor=\color{backcolour},   
    commentstyle=\color{codegreen},
    keywordstyle=\color{magenta},
    numberstyle=\tiny\color{codegray},
    stringstyle=\color{codepurple},
    basicstyle=\ttfamily\footnotesize,
    breakatwhitespace=false,         
    breaklines=true,                 
    captionpos=b,                    
    keepspaces=true,                 
    numbers=left,                    
    numbersep=5pt,                  
    showspaces=false,                
    showstringspaces=false,
    showtabs=false,                  
    tabsize=2
}

\lstdefinestyle{yaml}{
    basicstyle=\ttfamily\small,
    keywordstyle=\color{blue},
    breaklines=true,
    showstringspaces=false,
    identifierstyle=\color{black},
    stringstyle=\color{red},
    commentstyle=\color{gray},
}

\lstdefinelanguage{yaml}{
    morekeywords={true,false,null,y,n},
    sensitive=false,
    morecomment=[l]{\#},
    morestring=[b]",
    morestring=[b]',
    style=yaml
}

\begin{document}
\title{Auto3DSeg for Brain Tumor Segmentation from 3D MRI in BraTS 2023 Challenge.}
%
%

\author{Andriy Myronenko \and 
Dong Yang \and
Yufan He \and
Daguang Xu 
}

\authorrunning{A. Myronenko et al.}
%
\institute{NVIDIA, Santa Clara, CA \\
\email{\{amyronenko,dongy,yufanh,daguangx\}@nvidia.com}}

\maketitle

\begin{abstract}

 In this work, we describe our solution to the BraTS 2023 cluster of challenges using Auto3DSeg\footnote{https://monai.io/apps/auto3dseg} from MONAI\footnote{https://github.com/Project-MONAI/MONAI}. We participated in all 5 segmentation challenges, and achieved the 1st place results in three of them: Brain Metastasis, Brain Meningioma, BraTS-Africa challenges, and  the 2nd place results in the remaining two: Adult and Pediatic Glioma challenges.   

\end{abstract}

\keywords{Auto3DSeg  \and MONAI \and Segmentation \and NVAUTO team.}

\section{Introduction}

Multimodal Brain tumor segmentation challenge (BraTS)  has established itself as one of the largest MICCAI challenges, where researchers can develop and evaluate their solutions to 3D brain MRI tumor segmentation~\cite{brats21,BratsAll2018,brats_karargyris2023,Brats1}. BraTS 2023 follows the previous years setup with more data variability and notable new sub-challenges to segment various tumor types in various cohorts. BraTS 2023 segmentation challenges includes the following sub-challenges: Adult Glioma segmentation (same as BraTS 2021), Meningioma segmentation~\cite{labella2023asnrmiccai}, Sub-Sahara-Africa Glioma segmentation (BraTS-Africa)~\cite{adewole2023brain}, Pediatric tumor segmentation~\cite{kazerooni2023brain}, and  Metastasis segmentation~\cite{moawad2023brain} from brain 3D MRI.

Magnetic Resonance Imaging (MRI) is a key diagnostic tool for brain tumor analysis, monitoring and surgery planning. Usually, several complimentary 3D MRI modalities are acquired - such as T1, T1 with contrast agent (T1c), T2 and Fluid Attenuation Inversion Recover (FLAIR) - to emphasize different tissue properties and areas of tumor spread.  For example the contrast agent, usually gadolinium, emphasizes hyperactive tumor subregions in T1c MRI modality.  

\begin{figure}[t] 
 	\centering
 	\includegraphics[clip=true, trim=0pt 0pt 0pt 0pt, width=0.99\textwidth]{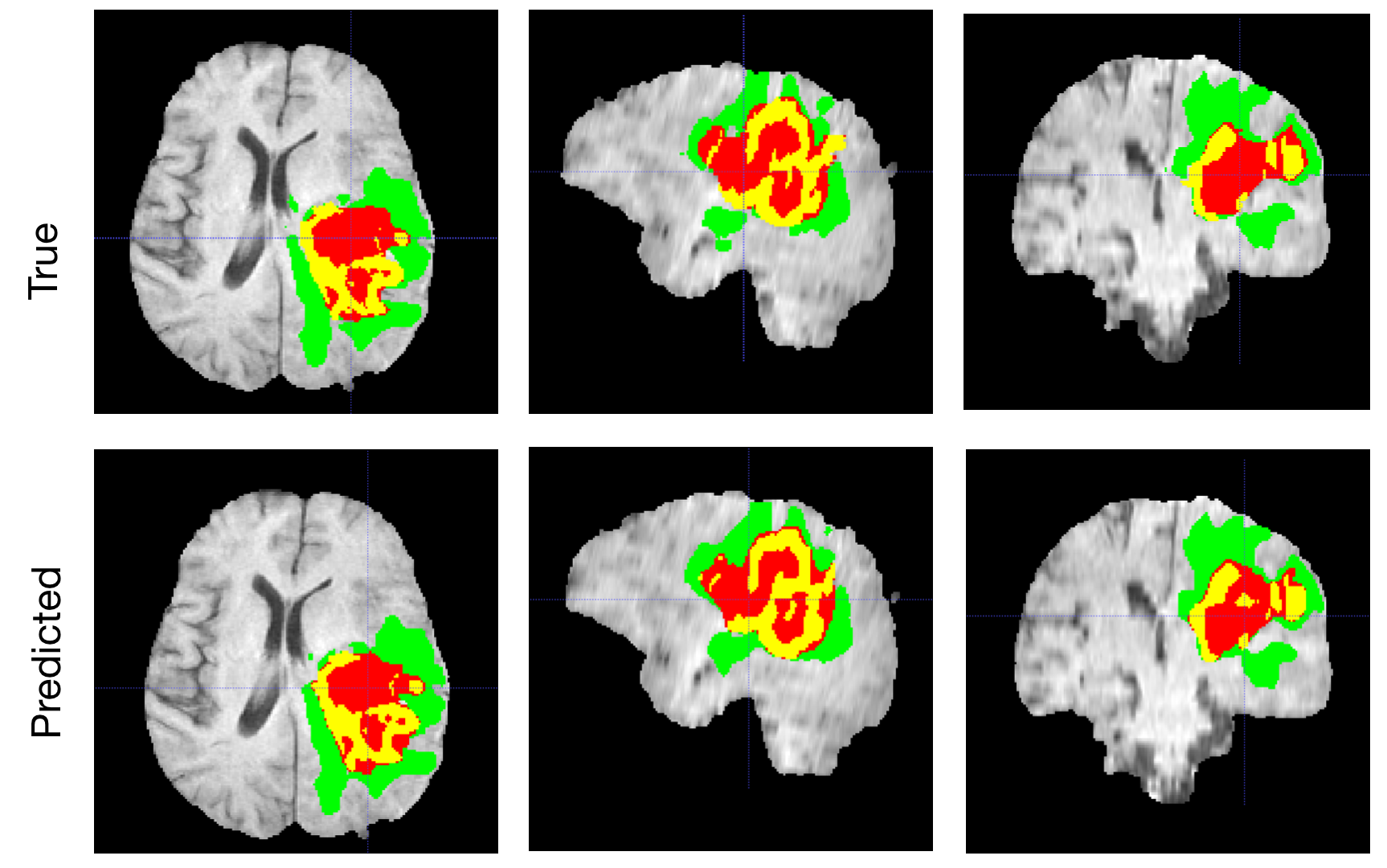}
 	\caption{A typical segmentation example with true and predicted labels overlaid over T1c MRI axial, sagittal and coronal slices.  The whole tumor (WT) class includes all visible labels (a union of green, yellow and red labels), the tumor core (TC) class is a union of red and yellow, and the enhancing tumor core (ET) class is shown in yellow (a hyperactive tumor part). }
 	\label{fig:seg}
 \end{figure}

BraTS aims to evaluate state-of-the-art methods for the segmentation of brain tumors by providing a 3D MRI dataset with ground truth tumor segmentation labels annotated by physicians~\cite{BratsAll2018,Brats1,Brats2,Brats3,Brats4}. BraTS training cases include four 3D MRI modalities (T1, T1c, T2 and FLAIR) rigidly aligned, resampled to 1x1x1 mm isotropic resolution and skull-stripped. The input image size is 240x240x155. The data were collected from multiple institutions, using various MRI scanners. Annotations include 3 tumor subregions: the enhancing tumor, the peritumoral edema, and the necrotic and non-enhancing tumor core.  The annotations were combined into 3 nested subregions: whole tumor (WT), tumor core (TC) and enhancing tumor (ET), as shown in Figure~\ref{fig:seg}.

\section{Methods}

We describe a solution that is automated and is easy to run even for non-expert users.  We use Auto3DSeg  utilizes open source components in MONAI~\cite{monai}, offering both beginner and advanced researchers the means to effectively develop and deploy high-performing segmentation algorithms.  The minimal user input to run Auto3DSeg for BraTS23 datasets, is 

\begin{lstlisting}[language=bash]
#!/bin/bash
python -m monai.apps.auto3dseg AutoRunner run \
    --input="./input.yaml" --algos=segresnet
\end{lstlisting}

where a user provided input config (input.yaml) includes several lines:

\begin{lstlisting}[language=yaml]
# This is the YAML file "input.yaml"
modality: MRI
datalist: "./dataset.json"
dataroot: "/data/brats23"

class_names:
- { "name": "wt", "index": [1,2,3] }
- { "name": "tc", "index": [1,3] }
- { "name": "et", "index": [3] }
sigmoid : true
      
\end{lstlisting}

When running this command,  Auto3DSeg will analyse the dataset, generate hyperparameter configs for several supported algorithms, trains them, and produce inference and ensembling.  The system will automatically scale to all available GPUs.

The 3 minimum user options (in input.yaml) are data modality (MRI in this case), location of the downloaded Brats23 dataset (dataroot), and the list of input filenames with an associated fold number (dataset.json). We generate the 5-fold split assignments randomly for each of the segmentation sub-challenges (since each of the BraTS 2023 sub-challenges share the same data structure).  

Since BraTS defines its specific label mapping (from integer class labels to 3 tumor subregions) we have to define it in the config,  and since these subregions are overlapping, we use "sigmoid: true" to indicate multi-label segmentation, where the final activation is sigmoid (instead of the default softmax).

Currently the supported  3D segmentation algorithms are SegResNet~\cite{myronenko20183d}, DiNTS~\cite{he2021dints} and SwinUNETR~\cite{hatamizadeh2021swin,tang2022self} with their unique training recipes. Here we used only the SegResNet algorithms, and trained it using 5-fold crooss validation.

The simplicity of Auto3DSeg is a very minimal user input, which allows even non-expert users to achieve a great baseline performance. The system will take care of most of the work to anaylyse, configure and optimally utilize the available GPU  resources.  And for expert users, there are many configurations options that can manually provided to override the automatic values, for better performance tuning.


\subsection{SegResNet}

\begin{figure}[t]
    \centering
    \includegraphics[width=0.8\textwidth]{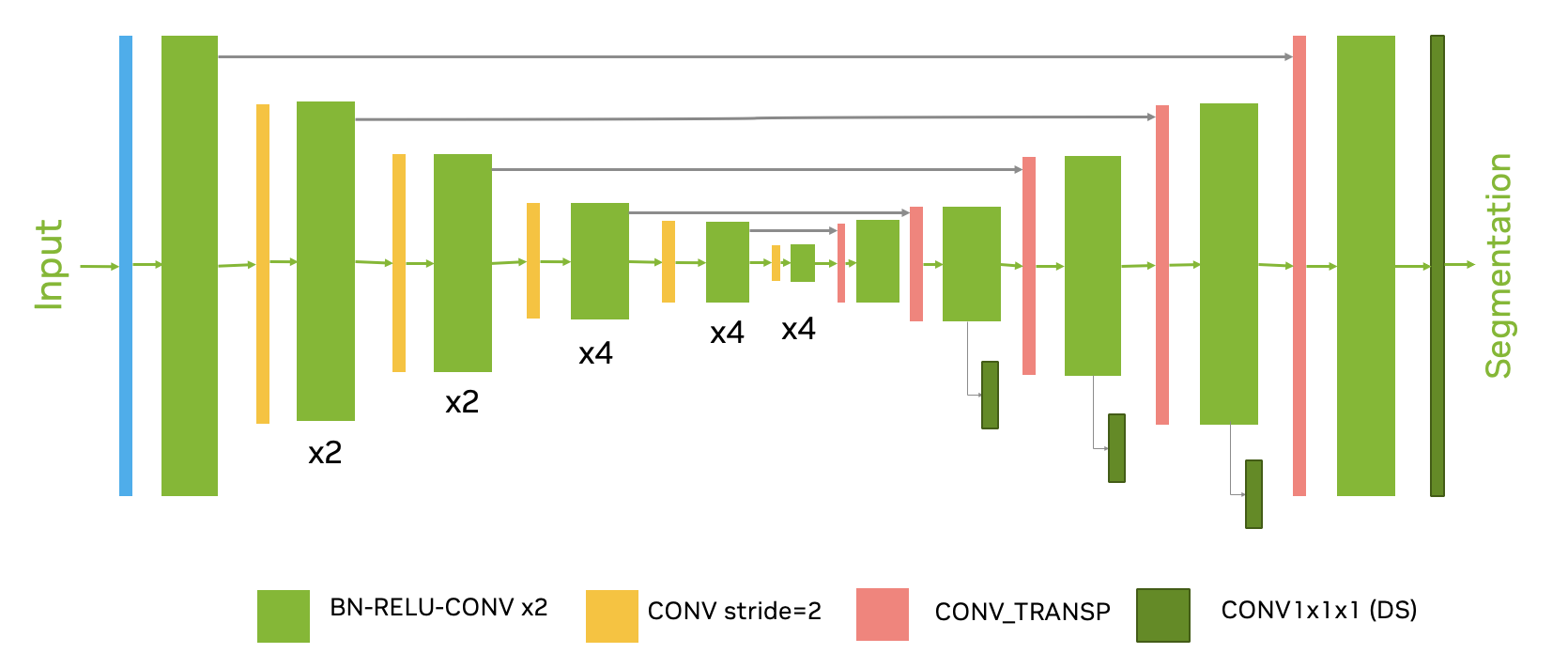}

    \caption{SegResNet network configuration. The network uses repeated ResNet blocks with batch normalization and deep supervision}
    \label{fig:net}
\end{figure}

SegResNet\footnote{https://docs.monai.io/en/stable/networks.html} is an encode-decoder based semantic segmentation network based on~\cite{myronenko20183d} It is a U-net based convolutional neural network (see Figure~\ref{fig:net}).

We use spatial augmentation including random affine and flip in all axes,random intensity scale, shift, noise and blurring.  We use the combined Dice + Focal loss. 

The same loss is summed over all deep-supervision sublevels:

\begin{equation}
Loss= \sum_{i=0}^{4} \frac{1}{2^{i}} Loss(pred,target^{\downarrow}) 
\end{equation}
where the weight $\frac{1}{2^{i}}$ is smaller for each sublevel (smaller image size) $i$. The target labels are downsized (if necessary) to match the corresponding output size using nearest neighbor interpolation

We use the AdamW optimizer with an initial learning rate of $2e^{-4}$ and decrease it to zero at the end of the final epoch using the Cosine annealing scheduler. We use batch size of 1 per GPU, We use weight decay regularization of $1e^{-5}$.

\subsection{Pre-processing}
We normalize input images to zero mean, unit standard deviation.

\subsection{Training}
We train the method on 8 GPUs 16GB NVIDIA V100  machine for 600 epochs.  We trained the models from scratch, with the  exception of BraTS-Africa dataset, which is very small. There, we initialized models from the checkpoints trained on the Glioma segmentation subtask (which was allowed by the competition).

Generally for all 5 segmentation challenges the training process was the same.  One exception was the Brain Metastasis challenge datasets, where the data from two sources (UCSF and Stanford) had a missing T2 MRI modality, with only one ground truth sub-region annotated - ET (TC and WT were missing).  To be able to use such data, we added a channel Dropout layer (prob=0.5) on T2 input channel. The dropout simulates channel to be all zeros, and help to make network robust to the cases when this channel is missing.  For the missing ground truth labels, we computed the loss only over the available ground truth classes (e.g. only for ET). Since we use a multi-label formulation (with the final sigmoid activation), we can easily compute loss only over the available classes (in-contrast to softmax acivation).  This was something unique only to the Metastasis dataset.

In a few cases, we re-trained (repeated training from the previous checkpoint) again, which showed a very slight validation performance improvement.  In retrospect, training longer (for more epochs) should achieve the same effect.  During 5-fold training we maintained not only the best average checkpoint, but also the best 3 sub-region checkpoints (ET, TC, WT). We observed that e.g. training twice (with a random initialization) will result in the same average dice, but individual dice scores per ET, TC, WT may differ between the train runs.

\subsection{Ensembling}

Our final result is an ensemble of 15 models (3 models from each fold) per each sub-region. Since during training we observed that the best average checkpoint may not be the best in all 3 sub-regions, we decided to ensemble models for each subregion (ET, TC, WT) separately using best sub-region checkpoints. Such approach, in the worst case, may require a total of 45 models, however in practice in most cases the best average checkpoint had also the best performance for all 3 or 2 of the sub-regions, so the total number of different model checkpoints used in the ensembling was about 25.   
 
\section{Results}

Based on our 5-fold data split,  average Dice (average of 3 sub-regions) for 5-folds cross-validation results of SegResNet model is shown in Table~\ref{tab:result}.  

\begin{table}[h!]
    \centering
    \begin{tabular}{ | c | c | c | c | c | c | c |}
        \hline
        & {\textbf{Fold 1}} & {\textbf{Fold 2}} & {\textbf{Fold 3}} & {\textbf{Fold 4}} & {\textbf{Fold 5}} & {\textbf{Average}} \\
        \hline
        Adult Glioma & 0.8076 & 0.925	 & 0.9211 &	0.9188	& 0.9205 & 0.9196 \\
        Meningioma & 0.9398	& 0.9403	& 0.939	& 0.9321	& 0.9251	& 0.93526 \\
        BraTS-Africa & 0.9144	& 0.8332	& 0.9005	& 0.7199	& 0.8322	& 0.84004 \\
        Pediatric & 0.8051	& 0.7844 &	0.794	& 0.7099 &	0.7393 &	0.76654 \\

        \hline
    \end{tabular}
    \caption{Avg Dice metric using internal 5-fold cross-validation.}
    \label{tab:result}
\end{table}

We have submitted 5  docker containers (for the 5 segmentation challenges), that vary in the model checkpoint weights.  The organizers ran our method on the hidden testing set, and returned back the metrics shown in Tables~\ref{tab:result_task1},~\ref{tab:result_task2},~\ref{tab:result_task3},~\ref{tab:result_task4},~\ref{tab:result_task5}, which we discuss below.

\begin{table}[h!]
    \centering
    \begin{tabular}{ | c | c | c | c | c |}
        \hline
        & {\textbf{Dice}} & {\textbf{HD95}} & {\textbf{Sensitivity}} & {\textbf{Specificity}}  \\
        \hline
        ET & 0.8076$\pm$0.25 & 40.88$\pm$94.81 & 0.8874$\pm$0.20 & 0.9997$\pm$0.0003  \\
        TC & 0.8278$\pm$0.27 & 38.85$\pm$91.82 & 0.9020$\pm$0.21 & 0.9996$\pm$0.0010	 \\
        WT & 0.8356$\pm$0.22 & 42.01$\pm$82.24 & 0.9357$\pm$0.09 & 0.9995$\pm$0.0007	 \\
        Avg & 0.8236 & 40.58 &	0.9083	& 0.9996  \\

        \hline
    \end{tabular}
    \caption{Task 1 Segmentation - Adult Glioma testing results (provided by the organizers).}
    \label{tab:result_task1}
\end{table}

\begin{table}[h!]
    \centering
    \begin{tabular}{ | c | c | c | c | c |}
        \hline
        & {\textbf{Dice}} & {\textbf{HD95}} & {\textbf{Sensitivity}} & {\textbf{Specificity}}  \\
        \hline
        ET & 0.7903$\pm$0.26 & 48.33$\pm$102.4 & 0.8738$\pm$0.21 & 0.9995$\pm$0.0004  \\
        TC & 0.8844$\pm$0.26 & 33.96$\pm$97.62 & 0.8785$\pm$0.22 & 0.9999$\pm$0.0003	 \\
        WT & 0.9144$\pm$0.15 & 21.49$\pm$57.63 & 0.9479$\pm$0.03 & 0.9994$\pm$0.0004	 \\
        Avg & 0.8497 & 35.59 &	0.9001	& 0.9995  \\

        \hline
    \end{tabular}
    \caption{Task 2 Segmentation - BraTS-Africa testing results (provided by the organizers).}
    \label{tab:result_task2}
\end{table}

Adult Glioma results based on the hidden test set (Table~\ref{tab:result_task1}), are much lower on average compared to our internal validation, indicating the test set is substantially different from  our 5-fold random split validation subsets (except for the Fold-1, which matched the performance).  This result achieves the overall 2nd rank in the Adult Glioma segmentation challenge. 

BraTS-Africa results based on the hidden test set are shown in Table~\ref{tab:result_task2}.  The BraTS-Africa dataset was relatively small, but unique due its population. First, most lesions were late stage, as the disease is detected and diagnosed at relatively late stage, unfortunately, in some areas of Africa. Secondly, the MRI scanners are often outdated, which results in lower image quality. Nevertheless, our method achieves the 1st rank in the BraTS-Africa segmentation challenge. 

Meningioma  results are  shown in Table~\ref{tab:result_task3}. It was the first time, brain meningioma data was used in BraTS, and the organizers assembled a large dataset. Our method performance is better than for Adult Glioma segmentation (which had a similar dataset size), indicating that perhaps there is less variability in Meningioma tumors, or that our 5-fold cross validation matched the hidden test set better. Our method achieved the 1st rank in the brain Meningioma segmentation challenge. 

\begin{table}[h!]
    \centering
    \begin{tabular}{ | c | c | c | c | c |}
        \hline
        & {\textbf{Dice}} & {\textbf{HD95}} & {\textbf{Sensitivity}} & {\textbf{Specificity}}  \\
        \hline
        ET & 0.8985$\pm$0.18 & 23.86$\pm$68.45 & 0.9364$\pm$0.12 & 0.9999$\pm$0.0001  \\
        TC & 0.9035$\pm$0.17 & 21.82$\pm$64.66 & 0.9412$\pm$0.11 & 0.9999$\pm$0.0004	 \\
        WT & 0.8709$\pm$0.19 & 31.39$\pm$71.80 & 0.9317$\pm$0.12 & 0.9998$\pm$0.0002	 \\
        Avg & 0.8910 & 25.69 &	0.9364	& 0.9999  \\

        \hline
    \end{tabular}
    \caption{Task 3 Segmentation - Meningioma testing results (provided by the organizers).}
    \label{tab:result_task3}
\end{table}

Metastasis results are shown in  Table~\ref{tab:result_task4}.  Again it was the first time, brain metastasis (secondary tumor) data were used in BraTS. This data had the most variability, with some institutional sources had incomplete data/labels. Metastatic tumors are especially challenging since there could be many of them present in the brain, and some of them could be very small and barely visible in the 3D MRI. For these reasons, the average dice metric results are much lower than the corresponding metric for Glioma and Meningioma data.  Nevertheless, our method achieved the 1 rank in the brain Metastasis segmentation challenge. 

\begin{table}[h!]
    \centering
    \begin{tabular}{ | c | c | c | c | c |}
        \hline
        & {\textbf{Dice}} & {\textbf{HD95}} & {\textbf{Sensitivity}} & {\textbf{Specificity}}  \\
        \hline
        ET & 0.6034$\pm$0.23 & 90.64$\pm$96.88 & 0.7541$\pm$0.23 & 0.9999$\pm$0.0001  \\
        TC & 0.6513$\pm$0.23 & 83.11$\pm$93.26 & 0.7952$\pm$0.24 & 0.9999$\pm$0.0001	 \\
        WT & 0.6210$\pm$0.23 & 88.14$\pm$92.74 & 0.7700$\pm$0.24 & 0.9999$\pm$0.0004	 \\
        Avg & 0.6253 & 87.30 &	0.7731	& 0.9998  \\

        \hline
    \end{tabular}
    \caption{Task 4 Segmentation - Metastasis testing results (provided by the organizers).}
    \label{tab:result_task4}
\end{table}

Finally, the Pediatric Glioma results are shown in Table~\ref{tab:result_task5}. This data type also had its first ever appearance at BraTS. The pediatric glioma data is very rare, compared to 
the adult gliomas, hence the dataset size was small. Our method achieved the 2nd rank in the Pediatric glioma segmentation challenge.

\begin{table}[h!]
    \centering
    \begin{tabular}{ | c | c | c | c | c |}
        \hline
        & {\textbf{Dice}} & {\textbf{HD95}} & {\textbf{Sensitivity}} & {\textbf{Specificity}}  \\
        \hline
        ET & 0.5458$\pm$0.36 & 115.31$\pm$144.03 & 0.7740$\pm$0.20 & 0.9994$\pm$0.0010  \\
        TC & 0.7815$\pm$0.19 & 27.09$\pm$72.11 & 0.7376$\pm$0.13 & 0.9997$\pm$0.0002	 \\
        WT & 0.8361$\pm$0.16 & 18.104$\pm$62.77 & 0.8041$\pm$0.09 & 0.9997$\pm$0.0003	 \\
        Avg & 0.7211 & 53.48 &	0.7719	& 0.9996  \\

        \hline
    \end{tabular}
    \caption{Task 5 Segmentation - Pediatric testing results (provided by the organizers).}
    \label{tab:result_task5}
\end{table}

\section{Discussion and Conclusion}
We participated in 5 segmentation challenges of BraTS23 using Auto3DSeg from MONAI. We customized the ensembling workflow (to ensemble each sub-region), and the training for brain metastasis (to account for missing modalities).  We achieved a strong performance, with the 1st place in the 3 (out of 5) challenges, and the 2nd place in the remaining two challenges.


%
%
\bibliographystyle{splncs04}
\bibliography{paper}

\end{document}